\documentclass{article}

\usepackage[preprint]{neurips_2026}

\usepackage[utf8]{inputenc} 
\usepackage[T1]{fontenc}    
\usepackage{url}            
\usepackage{booktabs}       
\usepackage{amsfonts}       
\usepackage{hyperref}       
\usepackage{nicefrac}       
\usepackage{microtype}      
\usepackage[table]{xcolor}  

\usepackage{amsmath}
\usepackage{amssymb}
\usepackage{multirow}
\usepackage{placeins}
\usepackage{array}
\usepackage{algorithm}
\usepackage{algpseudocode}
\usepackage{graphicx}
\usepackage{caption}
\usepackage{makecell}
\usepackage{wrapfig}
\usepackage{fontawesome}
\usepackage{tabularx}

\definecolor{routing}{HTML}{E6550D}
\colorlet{rankFirst}{routing!70}
\newcommand{\legendbox}[1]{\begingroup\setlength{\fboxsep}{0pt}\colorbox{#1}{\hspace{0.9em}\rule{0pt}{0.9em}}\endgroup}

\title{Parametric Memory Decoding for Zero-Shot Routing in LoRA-Based External Parametric Memory}
\author{
\textbf{Fengxian Ji}\textsuperscript{1}\thanks{Equal contribution.},
\textbf{Zhuohan Xie}\textsuperscript{1}\footnotemark[1],
\textbf{Jingpu Yang}\textsuperscript{2}\footnotemark[1],
\textbf{Fan Zhang}\textsuperscript{1},
\textbf{Zirui Song}\textsuperscript{1}, \\
\textbf{Xiuying Chen}\textsuperscript{1}\thanks{Corresponding author.} \\
$^{1}$MBZUAI, United Arab Emirates\\
$^{2}$Institute of Automation, Chinese Academy of Sciences, China\\
\texttt{\{fengxian.ji, Zhuohan Xie, zirui.song
\}@mbzuai.ac.ae}\\
\texttt{\{Fan Zhang, xiuying.chen\}@mbzuai.ac.ae, jingpuyang290@gmail.com}\\
}

\begin{document}
\maketitle
\begin{abstract}
With the rise of parametric memory, LoRA-based External Parametric Memory (EPM) has emerged as a modular solution, but existing routing methods often introduce additional training, deployment, and maintenance overhead.
This raises a natural question: can a LoRA-based EPM bank be routed without maintaining an additional routing component?
However, existing zero-shot LoRA routing methods still face two problems under the EPM setting: (1) their evaluations are scattered across different task settings rather than organized around EPM access, and (2) their routing signals lack a unified perspective to guide systematic improvement.
To address these problems, we organize \textbf{\textit{PMD-Bench}}, covering document-level, domain-level knowledge, and task-skill, and propose \textbf{\textit{Parametric Memory Decoding}} (PMD), the first framework designed to systematically improve zero-shot LoRA routing by reframing it as decoding activations over external parametric memory.
Based on PMD, we further instantiate \textbf{\textit{PMDRouter}}, which scores each LoRA by its response magnitude from a single base-model prefill.
Experiments on PMD-Bench show that PMDRouter achieves the strongest internal-signal performance across multiple zero-shot routing settings. These results demonstrate the feasibility of zero-shot LoRA routing and suggest that PMD can serve as a general framework for improving zero-shot routing methods.
Sources: \faGithub~ \href{https://anonymous.4open.science/r/Parametric-Memory-Decoding-872A/}{Github}.
\end{abstract}

\section{Introduction}
\begin{figure*}[t]
\centering
\includegraphics[width=\textwidth,keepaspectratio]{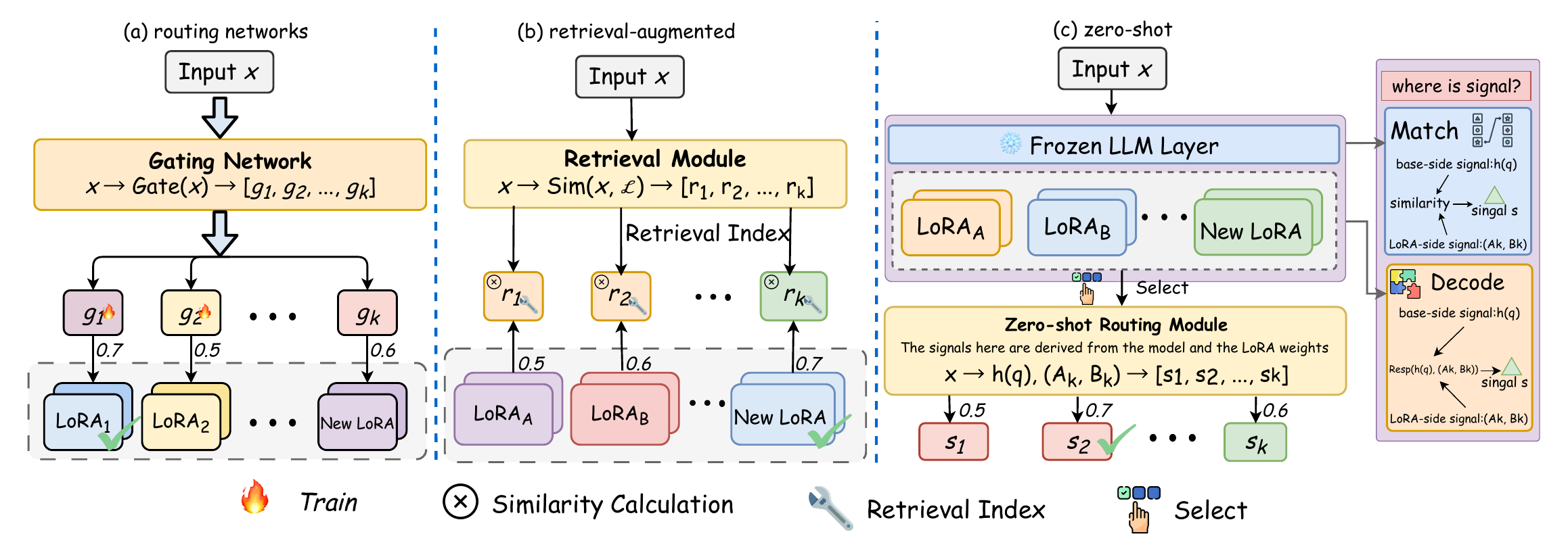}
\caption{
\textbf{Three LoRA routing paradigms}:
(a) learned routing networks,
(b) retrieval-augmented routing, and
(c) zero-shot LoRA routing, where scores are computed from the query-side signal \(h(q)\) and observable LoRA weights \((A_k,B_k)\).
PMD defines the routable response object as \(\rho_k(q)=\mathrm{Resp}(h(q),(A_k,B_k))\), and decodes it as \(s_k=D(\rho_k(q))\).
}
\label{fig:different_methods}
\end{figure*}

With the development of large language models, Large language model memory has attracted increasing attention~\citep{zhang2026memadapter,wei2025mlp,zhang2025survey,shazeer2017outrageously,fedus2022switch,yang2026zero}.
A recent form of model memory is parametric memory, which enables implicit and generalizable access to stored knowledge but is hard to update or modularize in monolithic models, motivating \emph{external parametric memory} (EPM) with independent parameter modules attached to a frozen backbone~\citep{back2026understandingloraknowledgememory,mallen2023not,lewis2020retrieval}.
Since EPM consists of multiple lightweight parameter modules, access becomes a routing problem: selecting which LoRA memory unit to activate for each query~\citep{bini2025memlora}.
As illustrated in Figure~\ref{fig:different_methods}(a,b), existing routing methods can be broadly divided into three categories by Lora routing methods, among which routing networks and retrieval-augmented routing rely on either learned routers or external representation spaces for adapter selection~\citep{chen2024llava,gao2025mola,zhao2024loraretriever,wang2025milora}.
While effective, these components add additional training, deployment, and maintenance overhead, motivating our question: Can a LoRA memory bank be accessed by a zero-shot policy?

As illustrated by the zero-shot paradigm in Figure~\ref{fig:different_methods} (c), this intrinsic access setting attempts to select LoRA modules that explore this direction by designing routing signals from the query, the backbone, and observable LoRA-side information. 
Representative methods such as Arrow, SEQR, HiLoRA, SpectR, and LAG derive routing signals from task vectors, adapter-side statistics, or internal model states~\citep{wang2022language,han2025hilora,cao2025memory}. 
These methods show that LoRA routing can be performed without an additional trained router, but two issues remain under the EPM setting: \textbf{Evaluation:} fragmented memory forms and access assumptions lead to inconsistent characterization across memory granularities; \textbf{Method:} routing signals are often constructed by searching over many query-side, LoRA-side, and scoring choices, yet can remain poorly separable for similar memory units.

These two issues point to two concrete problems in current zero-shot LoRA routing:
\textbf{Evaluation.}
Existing evaluations are not organized around a unified protocol of external parametric memory access. 
Many LoRA routing benchmarks are built around task-skill adapter banks, while others focus on document- or knowledge-oriented retrieval settings~\citep{zhao2024retrieval}. At the same time, terms such as training-free, data-free, and zero-shot are often used inconsistently~\citep{xia2025cross,lu2025little,lazier2025ac,fleshman2025spectr,zhang2026finreporting,zhang2026beyond}; here, zero-shot LoRA routing excludes routing-specific training data, retrieval indexes, and external representation spaces.
\textbf{Method.}
Existing zero-shot routers also lack a compact principle for constructing routing signals. In practice, they often search over many query-side representations, LoRA-side descriptors, scoring rules, layers, modules, and normalization choices~\citep{dhasade2026effective,fleshman2025seqr,han2025hilora}. While effective in some settings, these signals can be difficult to separate when candidate LoRA memories are trained on similar domains, related tasks, or overlapping documents.

To address these problems, we first organize \textbf{\textit{PMD-Bench}}, a multi-granularity benchmark for EPM access that covers \textsc{PaperQA} for document-level memory, \textsc{NQ-DomainLoRA} for domain-level knowledge memory, and \textsc{Task-LoRA} for task-skill memory.
We then propose \textbf{\textit{Parametric Memory Decoding}} (PMD), the first framework designed to systematically improve zero-shot LoRA routing by reframing it as decoding activations over external parametric memory.
Based on PMD, we instantiate \textbf{\textit{PMDRouter}}, which takes semantic activations from a single base-model prefill and scores each LoRA by its query-conditioned response magnitude.
Experiments on PMD-Bench show that PMDRouter achieves the strongest internal-signal performance across multiple zero-shot routing settings and stable gains across diverse memory granularities.
Overall, these results demonstrate the feasibility of zero-shot LoRA routing and suggest that PMD can serve as a general framework for improving zero-shot routing methods.

Our contributions are threefold:
(1) We organize \textbf{PMD-Bench}, the multi-granularity benchmark for external parametric memory routing. It covers document-level, domain-level knowledge, and task-skill memory with 35 memory-unit partitions in total (Section~\ref{subsec:PMD-BENCH}).
(2) We introduce \textbf{Parametric Memory Decoding} (PMD), which formulates zero-shot LoRA routing as decoding over external parametric memory rather than query-to-adapter matching. PMD reduces the estimated routing-design complexity by about \(40\times\) while improving routing accuracy (Sections~\ref{subsec:pmd_framework}, \ref{subsec:problem_complexity_analysis}).
(3) We propose \textbf{PMDRouter}, a parameter-free linear-response router that scores LoRAs by response energy from a single backbone prefill. PMDRouter achieves the strongest internal-signal zero-shot performance in most settings on PMD-Bench, with gains up to \(63.8\) points on Task-LoRA (Sections~\ref{sec:method}, \ref{subsec:Main_result}).

\section{Related Work}
\paragraph{External Parametric Memory}
Recent work increasingly views model parameters as memory for storing and accessing knowledge. Beyond factual knowledge, such parametric memory can also encode task skills, behavioral patterns, and domain preferences~\citep{zhang2024evaluating,wadhwa2024rags,wang2024wise,qiao2026memory,sprechmann2018memory,heinzerling2021language,valadarsky2017learning,zhang2026alpha,10.1117/12.3096291,10800533}. With the rise of PEFT methods such as adapters and LoRA, this memory is no longer confined to a monolithic set of model weights, but can instead be externalized into smaller, modular, and composable parameter units.

\paragraph{LoRA Routing}
Existing LoRA routing methods access a LoRA bank through different mechanisms, but under different assumptions. Some rely on trained routers, gating networks, or layer-/token-level expert allocation~\citep{chen2024llava,buehler2024x, araujo2024learning, gao2025mola,li2025dynmole,zhuang2025ld,sawada2020network,cai2021dynamic,muqeeth2024learning,huang2026hybrid}; others use task representations or external embedding spaces to retrieve or assign LoRA modules~\citep{kong2024lora,page2023multi,jin2026beyond,shukla2026qa,wu2024mixture,he2023routing,yang2025neural,li2024mixlora,prabhakar2025lora,cao2026taskspecificefficiencyanalysissmall}. While effective, these methods mix memory content with extra routing capacity, and therefore do not isolate whether the memory bank itself is accessible from internal query signals alone.
To isolate whether the memory bank is intrinsically decodable, one must focus on the zero-shot, zero-shot-based setting considered here, rather than related directions such as data-free LoRA transfer, query-adaptive fusion, LoRA-augmented generation, and LoRA reuse~\citep{chendoes,ponti2023combining,tian2025adapters,zhang2024dlp,li2025autolora,zhang2025unraveling,su2025tensorized,li2025loracoe,cao2026comol,li2026adafuse,ji2026servimage,cui2026textalign,ji2026finestate}.

\paragraph{zero-shot LoRA routing}
Zero-shot LoRA routing focuses on constructing adapter-selection signals from the query, the frozen backbone, and observable LoRA-side information. Representative methods include Arrow, SEQR, HiLoRA, SPECTR, and LAG, which use task vectors, adapter statistics, spectral descriptors, activation or textual proxies, and other model-internal signals~\citep{wang2022language,fleshman2025seqr,han2025hilora,fleshman2025spectr,fleshman2025lora,cao2025memory, jeong2026preference,kappiyathmaple,yang2026labguard}. While effective, these methods mostly follow static or proxy query adapter matching, which can be sensitive to design choices and poorly separable for semantically close LoRA memories. PMD instead formulates routing as decoding the query-conditioned response induced by each adapter.

\begin{figure*}[t]
\centering
\includegraphics[width=\textwidth,keepaspectratio]{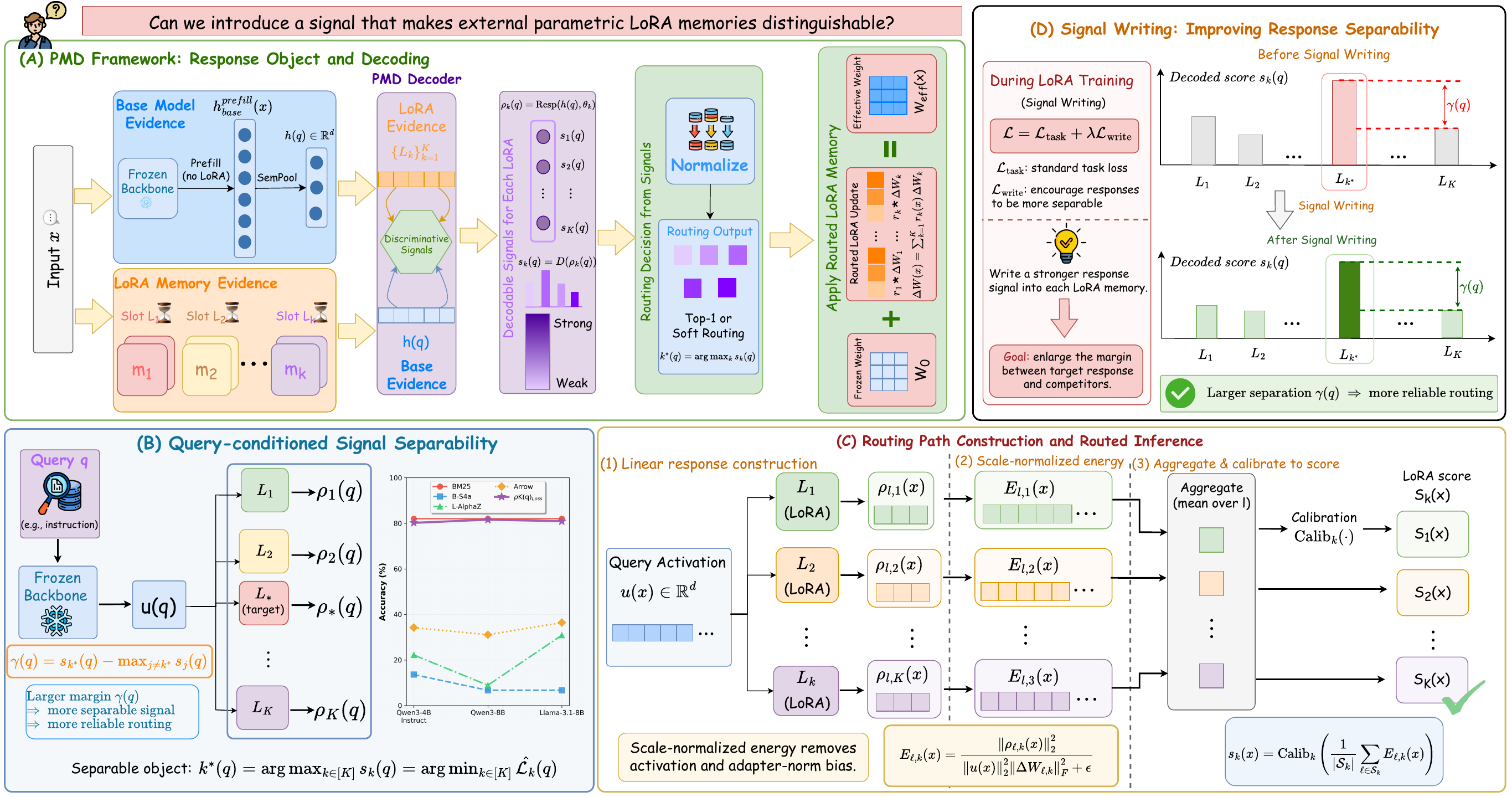}
\caption{
Overview of PMD and PMDRouter.
(A) PMD Framework: Response Object and Decoding;
(B) Query-conditioned Signal Separability;
(C) Routing Path Construction and Routed Inference;
(D) Signal Writing: Improving Response Separability.
}
\label{fig:framework}
\end{figure*}

\section{Zero-shot LoRA Routing Reformulation}
\label{sec:problem_formulation}
\subsection{Problem Reformulation}
\label{subsec:problem_formulation}
We study zero-shot LoRA routing for LoRA-based external parametric memory. Given a frozen backbone \(f_{\mathrm{base}}\) and a bank of post-attached LoRA adapters \(\mathcal{M}=\{L_k\}_{k=1}^K\), each adapter \(L_k\) acts as a parameterized memory unit that may encode domain knowledge, task skills, or their hybrid form. Given a query \(q\), the router must select which adapter to activate.
In this setting, the routing rule is constructed without learning a separate router: for each query, the query-dependent routing score is computed from frozen-backbone activations and observable LoRA weights, under a fixed scoring rule rather than an external retrieval representation.

Formally, let the \(k\)-th LoRA module be parameterized by low-rank factors \(\theta_k=(A_k,B_k)\), with update \(\Delta W_k=B_kA_k\). The query-side signal is obtained from the frozen backbone through a fixed parameter-free aggregation \(h(q)=\mathcal{A}(\{h_\ell^{\mathrm{base}}(t)\}_{t=1}^{T})\), where \(\mathcal{A}\) may denote last-token pooling, mean pooling, or another fixed aggregation over the backbone trajectory. We then define a parameter-free access score \(\mathcal{R}(h(q),A_k,B_k)\) and select the accessed memory slot by
\begin{equation}
k^*(q)
=
\arg\max_{k\in[K]}
\mathcal{R}\bigl(h(q),A_k,B_k\bigr).
\label{eq:ima_access_rule}
\end{equation}

\subsection{PMD-BENCH}
\label{subsec:PMD-BENCH}
To instantiate external parametric memory access in representative scenarios, we introduce PMD-Bench, a multi-granularity benchmark suite that measures whether LoRA-based memory banks are addressable across coarse-grained domain knowledge, fine-grained document knowledge, and task-oriented capability access, instantiated by NQ-DomainLoRA, PaperQA, and Task-LoRA, respectively.
Training data are converted into a unified two-turn \texttt{messages} format for LoRA training, while evaluation data are kept as independent QA or task instances for access selection and downstream scoring. Detailed dataset statistics and construction procedures are provided in Appendix~\ref{app: dataset}. Together, these benchmarks provide a compact but diverse suite for studying zero-shot access to external parametric memory across knowledge, document, and skill memory settings.

\section{METHODOLOGY}
\label{sec:method}
\subsection{Framework Overview}
\label{subsec:Framework_Overview}
Figure~\ref{fig:framework} summarizes our methodology. Under the zero-shot formulation in Section~\ref{subsec:problem_formulation}, PMD views backbone-and-LoRA-only routing as decoding query-conditioned responses rather than matching queries to static adapter descriptors. Each memory unit \(L_k\) induces a response object \(\rho_k(q)\), which is mapped to an access score and judged by its separability from competing responses.
PMDRouter instantiates PMD with a low-rank linear response and a calibrated energy decoder, requiring only one adapter-free backbone prefill and no trained gate, external retrieval, or per-LoRA trial generation. The same view also motivates an optional training-side signal-writing objective that strengthens the response signal used at inference time.

\subsection{PMD Framework}
\label{subsec:pmd_framework}
Section~\ref{subsec:problem_formulation} defines the admissible information for zero-shot LoRA routing: frozen backbone activations and observable LoRA weights. PMD specifies how to turn this information into a routing signal. 
PMD is a framework for constructing zero-shot LoRA routing signals by decomposing routing into response construction and response decoding:
\begin{equation}
s_k(q)
=
D\!\left(
\mathrm{Resp}\bigl(h(q),\theta_k\bigr)
\right),
\qquad
k^*(q)
=
\arg\max_{k\in[K]} s_k(q),
\label{eq:pmd_routing_score}
\end{equation}
where \(h(q)\) is the query representation produced by the frozen backbone, \(\theta_k=(A_k,B_k)\) denotes the observable parameters of the \(k\)-th LoRA memory unit, \(\mathrm{Resp}(\cdot,\cdot)\) constructs the query-conditioned response induced by this LoRA, and \(D\) decodes this response into a scalar routing score.

\paragraph{Separable response object.}
This formulation makes the routing object explicit. The router does not compare the query with a static adapter descriptor; instead, for each candidate LoRA \(L_k\), it computes an inference-time response signal:
\begin{equation}
\rho_k(q)
=
\mathrm{Resp}\bigl(h(q),\theta_k\bigr).
\label{eq:pmd_response_object}
\end{equation}
The response \(\rho_k(q)\) can be instantiated as a hidden-state perturbation, response energy, loss-related surrogate, or any other signal computed only from frozen backbone activations and LoRA weights. Thus, the routable object is the query-conditioned signal induced by the adapter, rather than the adapter identity or a static descriptor.
Given the response-derived score \(s_k(q)=D(\rho_k(q))\), routing succeeds when the target memory produces a score that is separated from competing memories. Let \(k^\star\) denote the target memory for query \(q\). We measure this separability by the PMD response margin \(\gamma(q)=s_{k^\star}(q)-\max_{j\neq k^\star}s_j(q)\). A larger \(\gamma(q)\) means that the target LoRA produces a more distinguishable response under the current query, making it easier to construct an effective decoder for routing.

\paragraph{Decoding pathway.}
Under this view, a concrete PMD router is determined by two design choices: the response constructor \(\mathrm{Resp}\), which decides what inference-time signal is read from each LoRA, and the decoder \(D\), which maps this signal to an access score. The resulting decoding pathway is
\begin{equation}
q
\rightarrow
h(q)
\rightarrow
\{\rho_k(q)\}_{k=1}^{K}
\rightarrow
\{s_k(q)\}_{k=1}^{K}
\rightarrow
r(q),
\label{eq:pmd_pathway}
\end{equation}
where \(r(q)\) is the final access pattern over the LoRA bank. In hard routing, \(r(q)\) selects one memory unit; in soft routing, it gives a distribution over candidate units.

\subsection{PMDRouter}
\label{subsection:PMDRouter}
PMDRouter instantiates the PMD solver in Section~\ref{subsec:pmd_framework} with a low-rank linear response operator and a scale-normalized energy decoder.
This gives a parameter-free routing rule that requires only one adapter-free backbone prefill and the frozen LoRA weights. 
Concretely, it realizes the abstract response object \(\rho_k(x)\) by applying each candidate LoRA update to a query-conditioned backbone activation, and then maps this response to an access score through scale-normalized response energy.
Directly probing each candidate LoRA with an estimated loss provides a strong separability signal, but it requires per-LoRA evaluation. PMDRouter serves as an efficient approximation to this loss-based signal by replacing per-LoRA loss probing with response-energy decoding; Appendix~\ref{app: Differnet_Paths_of_PMDRouter} implements and discusses several additional PMD routing paths under this framework.

\paragraph{Linear-response instantiation of \texorpdfstring{\(\mathrm{Resp}\)}{Resp}.}
Let \(f_{\mathrm{base}}\) be a frozen transformer with \(L\) layers, and let \(\mathcal{M}=\{L_k\}_{k=1}^{K}\) be a bank of LoRA memory units. For each adapted layer \(\ell\in\mathcal{S}_k\), the \(k\)-th LoRA provides a low-rank update \(\Delta W_{\ell,k}=B_{\ell,k}A_{\ell,k}\), where \(A_{\ell,k}\in\mathbb{R}^{r_\ell\times d}\) and \(B_{\ell,k}\in\mathbb{R}^{d\times r_\ell}\). Given an input \(x\), PMDRouter performs a single adapter-free prefill and extracts a routing activation \(u(x)=\mathrm{SemPool}(h^{\mathrm{base}}_{\ell_\star}(x))\in\mathbb{R}^{d}\), where \(\mathrm{SemPool}\) is a fixed pooling rule such as last-token pooling or span-level mean pooling.

Following the PMD definition in Eq.~\eqref{eq:pmd_response_object}, we instantiate the response object as the LoRA-induced first-order change on this activation:
\begin{equation}
\rho_{\ell,k}(x)
=
\mathrm{Resp}_{\mathrm{lin}}\bigl(u(x),\theta_{\ell,k}\bigr)
=
\Delta W_{\ell,k}u(x)
=
B_{\ell,k}A_{\ell,k}u(x).
\label{eq:pmdrouter-response}
\end{equation}
This response can be motivated from a local loss- or output-response view. For routing, however, we only need a score that preserves the relative ordering among candidate memories. Therefore, candidate-independent constants and higher-order terms in the local expansion are omitted, since they do not affect the ranking induced by the response score. 
This response is the basic object whose relative, scale-normalized strength will be decoded for routing.

\paragraph{Scale-normalized energy decoder.}
Given the responses \(\{\rho_{\ell,k}(x)\}_{\ell\in\mathcal{S}_k}\), PMDRouter decodes each response by a scale-normalized addressable energy:
\begin{equation}
E_{\ell,k}(x)
=
\frac{
\left\|
\rho_{\ell,k}(x)
\right\|_2^2
}{
\left\|
u(x)
\right\|_2^2
\left\|
\Delta W_{\ell,k}
\right\|_F^2
+
\epsilon
}
=
\frac{
\left\|
B_{\ell,k}A_{\ell,k}u(x)
\right\|_2^2
}{
\left\|
u(x)
\right\|_2^2
\left\|
B_{\ell,k}A_{\ell,k}
\right\|_F^2
+
\epsilon
}.
\label{eq:pmdrouter-energy}
\end{equation}
This score measures the relative response strength of the \(k\)-th memory unit after removing the scale effects of the input activation and LoRA update norm.

We aggregate the normalized energies across layers by mean pooling and obtain the final score \(\widetilde{E}_k(x)=\operatorname{Calib}_k(|\mathcal{S}_k|^{-1}\sum_{\ell\in\mathcal{S}_k}E_{\ell,k}(x))\). In our clean implementation, the calibration statistics are estimated from the training split and then fixed during evaluation; no test labels or trainable routing modules are used.

\paragraph{Decoding and routed inference.}
The calibrated addressable energy \(\widetilde{E}_k(x)\) is used as the PMD access score for the \(k\)-th LoRA memory unit, where the response object is the LoRA-induced linear response \(\rho_{\ell,k}(x)=B_{\ell,k}A_{\ell,k}u(x)\) and the decoder is the scale-normalized, calibrated response energy. 
In the hard-routing setting used in our main experiments, PMDRouter computes this score for all candidate LoRAs, selects the highest-scoring memory unit, and then runs the model with only the selected adapter activated. Thus, routing and generation are separated: the backbone is first used once without adapters to construct \(u(x)\), and the selected LoRA is used only in the final inference pass.

Concretely, PMDRouter instantiates the PMD score in Eq.~\eqref{eq:pmd_routing_score} as
\begin{equation}
s_k^{\mathrm{PMDRouter}}(x)
=
\widetilde{E}_k(x),
\qquad
\hat{k}(x)
=
\arg\max_{k\in[K]}
s_k^{\mathrm{PMDRouter}}(x).
\label{eq:pmdrouter_pmd_instantiation}
\end{equation}
This shows that PMDRouter realizes PMD by using query-conditioned LoRA responses as the routable object and calibrated addressable energy as the decoder.

\subsection{Training-Side Signal Writing}
\label{subsec:signal_writing}
PMD also suggests a training-side mechanism for improving future memory access. Since routing reliability depends on whether the target response is separated from competing responses, a LoRA bank becomes easier to access when the target unit exposes a stronger and more distinguishable response. When adapter training is controllable, we therefore augment the original task objective with a signal-writing loss:
\begin{equation}
\mathcal{L}
=
\mathcal{L}_{\mathrm{task}}
+
\lambda \mathcal{L}_{\mathrm{write}},
\qquad
\mathcal{L}_{\mathrm{write}}
=
\ell\bigl(g(\rho_k(x)), z_k\bigr).
\label{eq:signal_writing_objective}
\end{equation}

Here, \(\rho_k(x)\) denotes the query-conditioned response object in Eq.~\eqref{eq:pmd_response_object}, \(z_k\) is the response code assigned to memory unit \(k\), \(g(\cdot)\) extracts a response feature, and \(\ell(\cdot,\cdot)\) aligns it with the designated code. In \textsc{PMDRouter}, \(\rho_k(x)\) can be instantiated by the low-rank linear responses in Eq.~\eqref{eq:pmdrouter-response}, and \(g(\cdot)\) can be chosen to match the response-energy decoder in Eq.~\eqref{eq:pmdrouter-energy}. This objective does not train an additional router or add inference-time parameters; it shapes LoRA parameters so that the same PMD decoder reads a stronger response signal, thereby increasing the separation between the target score \(s_{k^\star}(q)\) and the strongest competing score \(\max_{j\neq k^\star}s_j(q)\).

\section{Experiments}
\label{sec:Experiments}
\subsection{Experimental Setup}
\label{sec:experimental_setup}
\paragraph{Baselines.}
We compare PMDRouter against four baseline families defined by the source of routing signal, covering Backbone-only, LoRA-only, joint Backbone-LoRA, and non-parametric retrieval settings under the same zero-shot constraints. 
(1) Backbone-only signal baselines use only the frozen base model's hidden states or logits. 
(2) LoRA-only signal baselines use only statistics derived from each LoRA module itself. 
(3) Joint Backbone-LoRA signal baselines combine backbone-side query representations with LoRA-side module information, yielding stronger zero-shot baselines without training an additional router; this family covers representative zero-shot and training-free routing methods such as Arrow~\citep{ostapenko2024towards}, SpectR~\citep{fleshman2025spectr}, and LAG~\citep{fleshman2025lora}. 
We also include (4) Text retrieval baselines with dense retrieval and BM25 to compare against non-zero-shot retrieval-based methods.
The full names of these methods are provided in Appendix~\ref{subapp:Baseline}.

\paragraph{Benchmark and evaluation sets.}
We evaluate all methods on the PMD-BENCH defined in Section~\ref{subsec:PMD-BENCH}: NQ-DomainLoRA, PaperQA, and Task-LoRA, which respectively represent domain-level knowledge memory, local document memory, and task-skill memory. For each benchmark, we train a bank of LoRA adapters on the corresponding training split and evaluate routing on the held-out evaluation split, where the router must select the most appropriate LoRA for each input under the zero-shot setting. We report routing accuracy (Acc). Detailed dataset construction and statistics are provided in Appendix~\ref{app: dataset}.

\begin{table*}[t]
    \centering
    \footnotesize
    \setlength{\tabcolsep}{1.8pt}
    \renewcommand{\arraystretch}{0.7}
    \begin{tabular}{lccccccccc} 
    \toprule
    \multirow{2}{*}{\raisebox{-1.5ex}{\textbf{Method}}}
    & \multicolumn{3}{c}{\textbf{PaperQA}}
    & \multicolumn{3}{c}{\textbf{NQ-DomainLoRA}}
    & \multicolumn{3}{c}{\textbf{TASK-Lora}} \\
    \cmidrule(lr){2-4}\cmidrule(lr){5-7}\cmidrule(lr){8-10} 
    & \makecell{\textbf{Llama-}\\\textbf{3.1-8B}}
    & \makecell{\textbf{Qwen3-4B-}\\\textbf{Instruct}}
    & \textbf{Qwen3-8B}
    & \makecell{\textbf{Llama-}\\\textbf{3.1-8B}}
    & \makecell{\textbf{Qwen3-4B-}\\\textbf{Instruct}}
    & \textbf{Qwen3-8B}
    & \makecell{\textbf{Llama-}\\\textbf{3.1-8B}}
    & \makecell{\textbf{Qwen3-4B-}\\\textbf{Instruct}}
    & \textbf{Qwen3-8B} \\
    
    \midrule
    \multicolumn{10}{c}{\textbf{Text retrieval}} \\ 
    \midrule
    \texttt{embedding}
    & 0.762 & 0.762 & 0.762
    & 0.397 & 0.397 & 0.397
    & 0.884 & 0.884 & 0.884 \\
    \texttt{bm25}
    & 0.820 & 0.820 & 0.820
    & 0.897 & 0.897 & 0.897
    & 0.735 & 0.735 & 0.735 \\
    
    \midrule
    \multicolumn{10}{c}{\textbf{Backbone--only Signals}} \\
    \midrule
    \texttt{B-LogitVar}
    & 0.120 & 0.082 & 0.078
    & 0.119 & 0.125 & 0.091
    & 0.263 & 0.288 & 0.170 \\
    \texttt{B-CtrL2}
    & 0.078 & 0.142 & 0.067
    & 0.099 & 0.144 & 0.103
    & 0.287 & 0.365 & 0.182 \\
    \texttt{B-S4a}
    & 0.067 & 0.136 & 0.067
    & 0.091 & 0.124 & 0.091
    & 0.126 & 0.256 & 0.178 \\
    
    \midrule
    \multicolumn{10}{c}{\textbf{LoRA--only Signals}} \\
    \midrule
    \texttt{L-AlphaZ}
    & 0.309 & 0.222 & 0.089
    & 0.186 & 0.211 & 0.120
    & 0.162 & 0.406 & 0.285 \\
    \texttt{L-HidL2-L4Z}
    & 0.269 & 0.171 & 0.080
    & 0.159 & 0.122 & 0.097
    & 0.179 & 0.377 & 0.180 \\
    \texttt{L-HidL2-LastMed}
    & 0.122 & 0.073 & 0.069
    & 0.147 & 0.212 & 0.084
    & 0.282 & 0.378 & 0.178 \\
    
    \midrule
    \multicolumn{10}{c}{\textbf{Joint Backbone--LoRA Signals}} \\
    \midrule
    \texttt{J-Global}
    & 0.062 & 0.096 & 0.138
    & 0.116 & 0.035 & 0.049
    & 0.357 & 0.301 & 0.091 \\
    \texttt{J-MLPGlobal}
    & 0.073 & 0.120 & 0.071
    & 0.083 & 0.083 & 0.083
    & 0.125 & 0.300 & 0.102 \\
    \texttt{J-WAllZ}
    & 0.080 & 0.080 & 0.064
    & 0.090 & 0.151 & 0.118
    & 0.256 & 0.219 & 0.066 \\
    \texttt{Arrow}
    & 0.364 & 0.342 & 0.311
    & 0.087 & 0.130 & 0.162
    & 0.158 & 0.139 & 0.108 \\
    \texttt{SPECTR}
    & 0.187 & 0.100 & 0.127
    & 0.601 & 0.570 & 0.474
    & 0.358 & 0.302 & 0.017 \\
    \texttt{LAG}
    & 0.440 & 0.598 & 0.636
    & 0.733 & \cellcolor{rankFirst}0.793 & 0.754
    & 0.315 & 0.326 & 0.217 \\
    \midrule    
    \texttt{PMDRouter(Ours)}
    & \cellcolor{rankFirst}0.600 & \cellcolor{rankFirst}0.613 & \cellcolor{rankFirst}0.658
    & \cellcolor{rankFirst}0.769 & 0.781 & \cellcolor{rankFirst}0.790
    & \cellcolor{rankFirst}0.663 & \cellcolor{rankFirst}0.897 & \cellcolor{rankFirst}0.923 \\
    \bottomrule
    \end{tabular}
    \caption{
    Main results on \textsc{PMD-Bench}. We report top-1 routing accuracy (\textbf{Acc}) across three memory granularities and three backbones. Shaded cells indicate the best zero-shot router in each column, excluding text-retrieval baselines. 
    }
    \label{tab:main-acc-merged}
\end{table*}

\subsection{Main result}
\label{subsec:Main_result}
Across PMD-Bench, PMDRouter is the strongest zero-shot router in most settings under the zero-shot protocol; see Table~\ref{tab:main-acc-merged}. 
Across PMD-Bench, PMDRouter is the strongest zero-shot router in most settings under the zero-shot protocol; see Table~\ref{tab:main-acc-merged}. 
On PaperQA, PMDRouter achieves accuracies of 0.600, 0.613, and 0.658 across the three backbones, consistently outperforming Backbone-only, LoRA-only, and joint Backbone-LoRA signal baselines. 
Its clearest advantage appears on Task-LoRA, where it reaches 0.663, 0.897, and 0.923, showing that task- and skill-oriented LoRA memories are highly decodable from query-conditioned parametric responses.
The pattern on NQ-DomainLoRA is more mixed. PMDRouter achieves the best zero-shot result on Llama-3.1-8B and Qwen3-8B, with accuracies of 0.769 and 0.790, but slightly trails LAG on Qwen3-4B-Instruct, with 0.781 compared to 0.793. This suggests that domain-level knowledge routing is more sensitive to backbone-specific response geometry and memory overlap than task-skill routing. 
GPT-5.2 judge scores, per-category accuracy, and failure cases are reported in Appendices~\ref{app:Judge_Scores}, \ref{app:additional_method_experimental_details}, and \ref{app:Cases Study}.

Comparison with text-retrieval baselines shows that parametric routing does not uniformly replace non-parametric retrieval.
On PaperQA and NQ-DomainLoRA, BM25 remains strong, especially because NQ-DomainLoRA documents are long and broad, which dilutes TF-IDF cosine signals, while BM25 benefits from exact lexical overlap with query keywords.
On Task-LoRA, PMDRouter is competitive with or stronger than retrieval baselines on the two Qwen backbones.
Thus, the main value of PMDRouter is to demonstrate effective zero-shot LoRA routing using only backbone and LoRA signals, rather than to dominate retrieval in every setting.

\begin{figure}[t]
    \centering
    \includegraphics[width=\linewidth,keepaspectratio]{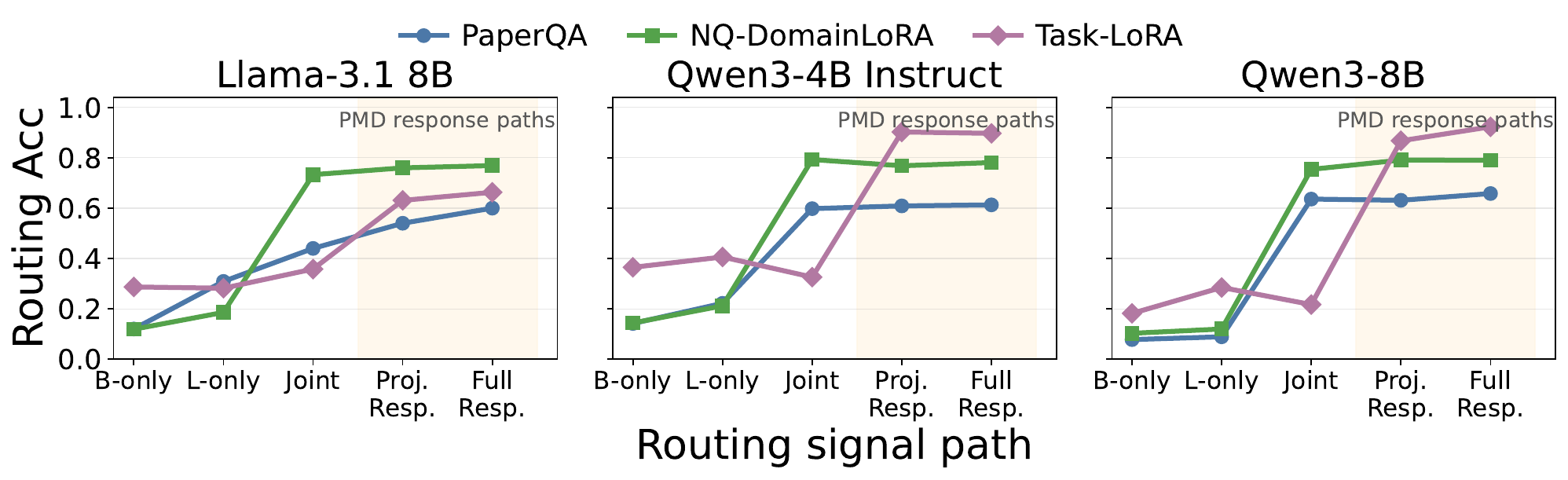}
    \caption{
    Routing accuracy across signal paths.
    Moving from single-side cues to joint and PMD response paths generally improves routing accuracy across backbones and benchmarks.
    }
    \label{fig:pmd-decodability-success}
\end{figure}

\subsection{Result Analysis}
\label{sec:Result_analysis}
\begin{figure}[t]
  \centering  \includegraphics[width=\linewidth,keepaspectratio]{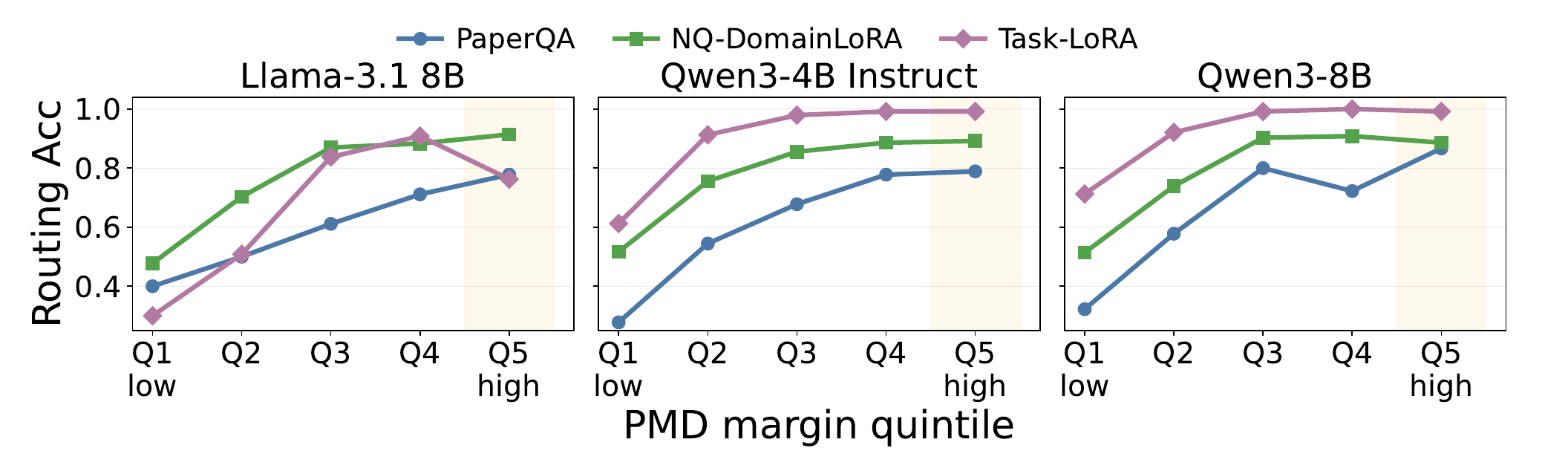}
    \caption{
    PMD response margin predicts routing success. Queries are grouped by normalized top-1/top-2 response margin; higher-margin groups generally achieve higher routing accuracy across backbones and benchmarks.
    }
  \label{fig:pmd-Addressability-success}
\end{figure}

We analyze PMDRouter from two complementary perspectives: whether response-based signal paths improve routing, and whether the PMD response margin explains routing behavior.
As shown in Figure~\ref{fig:pmd-decodability-success}, routing accuracy generally improves as the routing signal moves from single-side cues to joint and response-based paths.
Backbone-only and LoRA-only signals are consistently weak, while projection response and full response paths achieve stronger performance across backbones and benchmarks.
This supports the PMD view that routing should be constructed from query-conditioned interactions between the frozen-backbone signal and LoRA weights, rather than from either side alone.
We further examine the separability of the resulting PMD response signal.
For each query, we compute the normalized top-1/top-2 PMD response margin and group examples into five quintiles.
As shown in Figure~\ref{fig:pmd-Addressability-success}, routing accuracy generally increases from low-margin to high-margin groups across backbones and benchmarks.
This confirms that response separability is a key driver of memory addressability: when the target memory response is more clearly separated from competing responses, routing is more reliable.
The margin trend is strongest on PaperQA and Task-LoRA, which helps explain the large gains of PMDRouter in Table~\ref{tab:main-acc-merged}.
NQ-DomainLoRA also shows a clear increasing trend, although its high-margin groups tend to saturate, suggesting that domain-level knowledge routing is additionally affected by backbone-specific geometry and overlap among memory units.

\subsection{Improving Memory Addressability During Training}
\label{subsec:Training Improvement}

\begin{wraptable}{l}{0.48\columnwidth}
    \centering
    \scriptsize
    \renewcommand{\arraystretch}{0.8} 
    \resizebox{0.46\columnwidth}{!}{%
    \begin{tabular}{lcc}
    \toprule
    \textbf{Variant} & \textbf{Acc} & \textbf{NormGap} \\
    \midrule
    QMean-NoAux  & 0.629 & 0.466 \\
    QMean-Raw    & \cellcolor{rankFirst}0.689 & \cellcolor{rankFirst}0.554 \\
    QMean-Log    & 0.676 & 0.527 \\
    Last-Raw     & 0.631 & 0.479 \\
    Ortho        & 0.629 & 0.463 \\
    Mono         & 0.627 & 0.465 \\
    Base         & 0.613 & 0.463 \\
    \bottomrule
    \end{tabular}%
    }
    \caption{Training improvement for memory Addressability on PaperQA. We report routing accuracy (Acc) and normalized top-1/top-2 routing gap (NormGap) under the same inference-time router.  \textbf{Color:} \legendbox{rankFirst}\,1st.}
    \label{tab:training_improvement_singlecol}
\end{wraptable}
As discussed in Section~\ref{subsec:signal_writing}, although zero-shot routing already achieves competitive performance without a trained router, we further ask whether explicitly shaping the routing signal during LoRA training can improve memory access.
On \textsc{PaperQA}, we compare several training-time shaping strategies under the same inference-time router, \textsc{QMean-All-LogC}, which uses question-mean pooling, all-module response aggregation, and per-LoRA log-centered scoring; the corresponding loss formulations are provided in Appendix~\ref{app:training_objectives}.
All variants share the same task loss and add a routing-related auxiliary term, $\mathcal{L}=\mathcal{L}_{\text{task}}+\lambda \mathcal{L}_{\text{aux}}$. Here, QMean-All-LogC denotes the fully decoder-aligned shaping strategy, while QMean-All-NoAux and QMean-All-Raw are two controls that remove auxiliary shaping or replace log-transformed energy with raw energy, respectively.
As shown in Table~\ref{tab:training_improvement_singlecol}, QMean-All-LogC achieves the best result, reaching 69.78 Acc and 0.2637 NormGap on PaperQA, clearly outperforming Last-Raw (62.22 / 0.1549) and Ortho (61.78 / 0.1591). This indicates that training is most effective when it explicitly shapes the same response-energy signal family later used by the inference-time decoder. The pending controls, QMean-All-NoAux and QMean-All-Raw, will further test whether this gain comes from auxiliary shaping itself and from the use of a stable signal transform.

\begin{table*}[t]
\centering
\small
\setlength{\tabcolsep}{3.8pt}
\renewcommand{\arraystretch}{1.12}
\begin{tabular}{lccccccc}
\toprule
\multirow{2}{*}{\textbf{Formulation}}
& \multicolumn{3}{c}{\textbf{Design space}}
& \multicolumn{3}{c}{\textbf{Complexity}}
& \multirow{2}{*}{\textbf{Best Acc.}} \\
\cmidrule(lr){2-4}\cmidrule(lr){5-7}
& \textbf{Query}
& \textbf{Memory}
& \textbf{Decision}
& \textbf{Coupling}
& \textbf{Search Space}
& \textbf{Complexity}
&  \\
    \midrule
        Direct matching view & $\sim$9 & $\sim$11 & $\sim$56 & $\sim$21 & $\sim$56--$4.2\times 10^{5}$ & $\sim$1.00 & 0.636 \\
        PMD view             & $\sim$1 & $\sim$1 & $\sim$3 & $\sim$1 & $\sim$3 & $\sim$0.025 & 0.669 \\
    \bottomrule
\end{tabular}
\caption{
    Design-space complexity of direct matching and PMD.
    Lower values indicate simpler routing search; Best Acc. reports the top accuracy.
}
\label{tab:pmd_complexity_clean}
\end{table*}

\subsection{Problem Complexity Analysis}
\label{subsec:problem_complexity_analysis}
Beyond final routing accuracy, we analyze whether PMD also simplifies the structure of the zero-shot LoRA routing problem. Here, complexity refers not to runtime cost, but to the size and coupling of the design space required to construct a routing rule. As summarized in Table~\ref{tab:pmd_complexity_clean}, the direct matching view requires choices over query representations, memory descriptors, and decision rules, leading to a large and highly coupled search space. In contrast, PMD fixes the routing object as the query-conditioned response and leaves the solver to choose how this response is decoded.
Under the direct matching view, the router must jointly specify query-side signals, memory-side signals, and matching decisions, with additional coupling across these axes. This yields an estimated search space of \(\sim 56\)--\(4.2\times10^5\), coupling degree of \(\sim 21\), and normalized complexity of \(\sim 1.00\). Under the PMD view, the query and memory axes collapse into a response-centered formulation: the main remaining choices are reduced to the response instantiation and the decoder. This reduces the estimated search space to \(\sim 3\), coupling degree to \(\sim 1\), and normalized complexity to \(\sim 0.025\).
Despite this smaller design space, PMD also supports stronger routing performance on PaperQA, improving the best accuracy from \(0.636\) under the direct matching view to \(0.669\). This suggests that PMD is useful not only as a stronger routing formulation, but also as a way to rewrite zero-shot LoRA routing into a more compact and less coupled decoding problem.

\subsection{Ablation Study}
\label{sec:ablation}

\begin{wrapfigure}{r}{0.50\columnwidth}
    \centering
    \includegraphics[width=0.50\columnwidth]{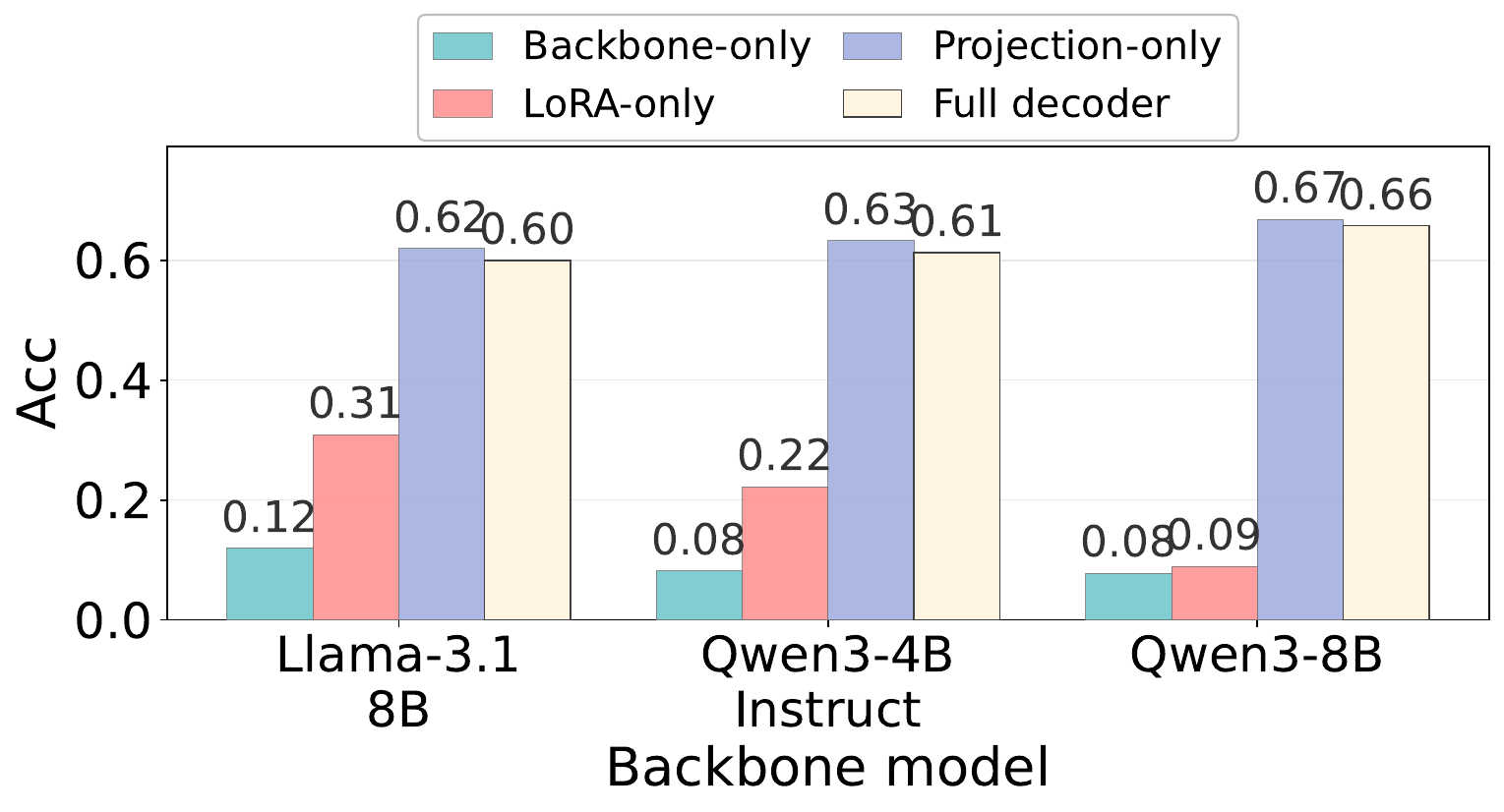}
    \caption{Ablation Study. Accuracy of different decoder adaptation settings across backbone models: Backbone only, LoRA only, Projection only, and Full decoder.}
    \label{fig:ablation_decoder_components}
\end{wrapfigure}

To clarify PMDRouter's gains, we compare four inference-time decoder settings on PaperQA: \textit{Backbone-only}, \textit{LoRA-only}, \textit{Projection-only}, and the full decoder. As shown in Figure~\ref{fig:ablation_decoder_components}, \textit{Backbone-only} and \textit{LoRA-only} are consistently weaker across all three backbones, showing that neither query-side nor adapter-side signals alone are sufficient for reliable LoRA selection. By contrast, \textit{Projection-only} improves substantially on all three backbones and is the strongest variant across all backbones, suggesting that query-conditioned low-rank interaction already captures most of the useful routing signal.
The full decoder is slightly worse than \textit{Projection-only} across all three backbones, indicating that full write-back does not necessarily yield stronger routing separability. A plausible explanation is that \(A s(x)\) is closer to memory addressing, whereas \(B A s(x)\) introduces the adapter-specific metric defined by \(B\), which may bring generation-oriented scale biases.
Overall, this ablation shows that zero-shot LoRA routing relies on decodable query-conditioned interactions, with the low-rank addressing projection providing the most stable signal.

\section{Conclusion}
We study zero-shot LoRA routing for LoRA-based external parametric memory without an additional routing component. We organize PMD-Bench to evaluate this setting across document-level, domain-level, and task-skill memory, and propose PMD, which reframes routing as decoding activations over external parametric memory.
Based on PMD, PMDRouter scores each LoRA by its response magnitude from a single adapter-free backbone prefill. Experiments show that it achieves the strongest internal-signal performance across multiple zero-shot settings, while retrieval remains competitive in some knowledge-oriented cases.
Overall, these results demonstrate the feasibility of zero-shot LoRA routing and suggest that PMD can guide future improvements in zero-shot routing methods.


\nocite{*}
{\small
\bibliographystyle{abbrv}
\bibliography{custom}
}

\appendix
\section{Dataset}
\label{app: dataset}
We construct three benchmark datasets: NQ-DomainLoRA, PaperQA, and Task-LoRA. All training data are formatted into a unified two-turn \texttt{messages} schema for LoRA training, while evaluation data are kept as question--answer items for independent scoring. 
Table~\ref{tab:dataset_summary_examples} summarizes the overall scale of the three datasets and presents representative evaluation examples.

\begin{table*}[htp]
\centering
\scriptsize
\setlength{\tabcolsep}{2.2pt}
\renewcommand{\arraystretch}{0.68}
\resizebox{0.92\textwidth}{!}{%
\begin{tabular}{
p{0.10\textwidth}
p{0.11\textwidth}
p{0.08\textwidth}
p{0.08\textwidth}
p{0.08\textwidth}
p{0.20\textwidth}
p{0.18\textwidth}
}
\toprule
\textbf{Dataset}
& \textbf{Unit}
& \textbf{Train}
& \textbf{Eval/Test}
& \textbf{Task type}
& \textbf{Example input}
& \textbf{Example answer} \\
\midrule
NQ-DomainLoRA
& 360 domain documents
& 7,200 messages
& 1,800 QA
& Domain QA
& How does the narrative focus shift between the animated trilogy and \textit{Dragons: The Nine Realms}?
& The trilogy focuses on Hiccup's coming of age in a mythical Viking setting, whereas \textit{The Nine Realms} shifts the story about 1,300 years into the future. \\
\midrule
PaperQA
& 15 paper introductions
& 795 messages
& 450 QA
& Paper QA
& What framework integrates diffusion-based trajectory generation with MCTS?
& The proposed framework is Monte Carlo Tree Diffusion (MCTD). \\
\midrule
Task-LoRA
& 8 task families
& 4,800 messages
& 1,200 messages
& NLI
& Determine the relationship between the premise and hypothesis. Return one label: entailment or not\_entailment.
& entailment \\
\bottomrule
\end{tabular}%
}
\caption{
Dataset statistics and representative evaluation/test examples for the three benchmark datasets.
}
\label{tab:dataset_summary_examples}
\end{table*}

\paragraph{NQ-DomainLoRA}.
NQ-DomainLoRA is designed to study \emph{knowledge-oriented} LoRA routing. We start from the Natural Questions training set and group examples by Wikipedia document using title- and URL-based domain rules. We keep 12 target domains: \textit{entertainment}, \textit{history}, \textit{sports}, \textit{geography}, \textit{music}, \textit{politics}, \textit{literature}, \textit{business}, \textit{science}, \textit{biology}, \textit{technology}, and \textit{food}. For each domain, we sample 30 documents, yielding 360 documents in total. Each document is paired with a Wikipedia plain-text extract truncated to at most 6,000 characters. For training, each document yields 20 messages: 1 read/memorize example, 15 factual QA examples, 2 summaries, and 2 rewrites. For evaluation, each document yields 5 QA items emphasizing cross-sentence synthesis and relation-level reasoning. The resulting dataset contains 7,200 training messages and 1,800 evaluation QA pairs.

\paragraph{PaperQA}.
PaperQA is designed to study \emph{local scientific knowledge memory}. We collect 15 papers from ICLR 2025, ICML 2025, and NeurIPS 2024, and extract their introduction sections from PDF or fallback sources. Each introduction is used to construct both training and evaluation data. The evaluation set contains 30 QA items per paper, organized into three difficulty levels: factual recall, contextual understanding, and logical/motivational inference. The training set contains 53 messages per paper: 1 reading example, 40 QA examples, 8 summaries, and 4 rewrites. In total, PaperQA contains 795 training messages and 450 evaluation QA pairs. The average introduction length is 5,131.67 characters, with a range from 2,165 to 14,298 characters.

\paragraph{Task-LoRA}.
Task-LoRA is designed to study \emph{task- and skill-oriented} LoRA routing. It covers 8 task families: BoolQ, edit/summarization, extractive QA, generative chain-of-thought, instruction following, multiple-choice QA, math chain-of-thought, and NLI. Raw data are collected from public task sources and converted into a unified \texttt{messages} format. For each task family, we shuffle with a fixed seed and take 600 samples for training and 150 for testing. The resulting dataset contains 4,800 training messages and 1,200 test messages in total.

For NQ-DomainLoRA and PaperQA, training files are stored as JSON objects whose keys are document or paper IDs and whose values are message lists; evaluation files are JSON objects mapping IDs to QA lists. For Task-LoRA, each task is stored as JSONL with fields including \texttt{id}, \texttt{task}, \texttt{messages}, and \texttt{meta}. This unified format allows all three datasets to be directly used for instruction tuning or LoRA training, while preserving evaluation annotations for separate scoring. As a reference compute estimate, training the nine LoRA-bank configurations used in our main experiments takes approximately 80 wall-clock hours on two NVIDIA RTX 6000 GPUs, and routing plus inference evaluation takes approximately 30 wall-clock hours on the same hardware.

\section{Derivations and Analysis}
\label{app: Derivations and Analysis}

\subsection{Baseline Methods}
\label{subapp:Baseline}

This section summarizes the baseline routing methods used in our experiments. 
For each method, we first describe its full name and signal source in text, and then give only the final routing signal.

\paragraph{\texttt{embedding}: TF-IDF Cosine Similarity Retrieval.}
Despite the name, this baseline uses sparse TF-IDF vectors rather than dense embeddings. 
It ranks adapters by the cosine similarity between the query text and the adapter- or paper-side text \(d_a\):
\begin{equation*}
s_a(x)
=
\cos\bigl(\mathrm{tfidf}(x),\mathrm{tfidf}(d_a)\bigr).
\end{equation*}

\paragraph{\texttt{bm25}: BM25 Lexical Retrieval.}
This baseline ranks adapters by lexical term matching between the query text and the adapter- or paper-side text \(d_a\). 
In our implementation, \(k_1=1.5\) and \(b=0.75\):
\begin{equation*}
s_a(x)
=
\sum_{t\in x}
\mathrm{IDF}(t)
\frac{
f(t,d_a)(k_1+1)
}{
f(t,d_a)+k_1\bigl(1-b+b|d_a|/\mathrm{avgdl}\bigr)
}.
\end{equation*}

\paragraph{\texttt{B-LogitVar}: Backbone-only Logit Delta Variance.}
This method uses the top-token logit perturbation obtained by projecting the LoRA-induced hidden perturbation through the LM head. 
Let \(\delta h_a=\sum_m\Delta W_{a,L,m}h^0_{L,m}(x)\) and \(\delta z_{a,t}=\langle W_t,\delta h_a\rangle\). 
The final signal is the weighted variance of the logit perturbation over the selected token set \(T\):
\begin{equation*}
s_a(x)
=
\sum_{t\in T}
\bar p_t
\left(
\delta z_{a,t}
-
\sum_{u\in T}\bar p_u\delta z_{a,u}
\right)^2 .
\end{equation*}

\paragraph{\texttt{B-CtrL2}: Backbone-only Centered Logit L2.}
This method uses the same logit perturbation as \texttt{B-LogitVar}, but scores each adapter by the distance of its perturbation from the adapter-wise center:
\begin{equation*}
s_a(x)
=
\left\|
\delta z_a
-
\frac{1}{|\mathcal A|}
\sum_{b\in\mathcal A}\delta z_b
\right\|_2 .
\end{equation*}

\paragraph{\texttt{B-S4a}: Backbone-only Positive Surprisal Alignment.}
This method aligns the positive part of the centered logit perturbation with the base-model surprisal:
\begin{equation*}
s_a(x)
=
\sum_{t\in T}
\left[
\delta z_{a,t}
-
\frac{1}{|\mathcal A|}
\sum_{b\in\mathcal A}\delta z_{b,t}
\right]_+
(-\log p_t).
\end{equation*}

\paragraph{\texttt{L-AlphaZ}: LoRA-only Alpha Norm with per-LoRA Z-score.}
This method uses the low-rank coordinate response of the LoRA \(A\) matrices to the backbone hidden states. 
Let \(\mu_a\) and \(\sigma_a\) denote the mean and standard deviation computed over all queries for the same LoRA \(a\):
\begin{equation*}
r_a(x)
=
\left(
\sum_m
\left\|
A_{a,L,m}h^0_{L,m}(x)
\right\|_2^2
\right)^{1/2},
\qquad
s_a(x)
=
\frac{r_a(x)-\mu_a}{\sigma_a}.
\end{equation*}

\paragraph{\texttt{L-HidL2-L4Z}: LoRA-only Hidden Centered L2 over Last-4 Layers with per-LoRA Z-score.}
This method uses centered hidden perturbations over the last four layers and takes the maximum layer score. 
Let \(\delta h_{a,\ell}=\sum_m\Delta W_{a,\ell,m}h^0_{\ell,m}(x)\), and let \(\mu_a,\sigma_a\) denote per-LoRA normalization statistics:
\begin{equation*}
r_a(x)
=
\max_{\ell\in\mathrm{last4}}
\left\|
\delta h_{a,\ell}
-
\frac{1}{|\mathcal A|}
\sum_{b\in\mathcal A}\delta h_{b,\ell}
\right\|_2,
\qquad
s_a(x)
=
\frac{r_a(x)-\mu_a}{\sigma_a}.
\end{equation*}

\paragraph{\texttt{L-HidL2-LastMed}: LoRA-only Last-layer Hidden Centered L2 with Median Center.}
This method uses the last-layer LoRA hidden perturbation centered by the adapter-wise median:
\begin{equation*}
s_a(x)
=
\left\|
\delta h_{a,L}
-
\operatorname{median}_{b\in\mathcal A}
\bigl(\delta h_{b,L}\bigr)
\right\|_2 .
\end{equation*}

\paragraph{\texttt{J-Global}: Joint Backbone-LoRA Global Geometry Score.}
This method measures the global cosine alignment between the LoRA perturbation and the base-model negative gradient. 
By default, it uses the last four layers, all prompt tokens, and the modules \(\mathrm{q\_proj}\), \(\mathrm{v\_proj}\), \(\mathrm{o\_proj}\), \(\mathrm{up\_proj}\), \(\mathrm{gate\_proj}\), and \(\mathrm{down\_proj}\):
\begin{equation*}
s_a(x)
=
\frac{
\sum_{\ell,m,t}
\left\langle
\Delta W_{a,\ell,m}h^a_{\ell,m,t}(x),
g_{\ell,m,t}(x,\hat y)
\right\rangle
}{
\sqrt{
\sum_{\ell,m,t}
\left\|
\Delta W_{a,\ell,m}h^a_{\ell,m,t}(x)
\right\|_2^2
}
\sqrt{
\sum_{\ell,m,t}
\left\|
g_{\ell,m,t}(x,\hat y)
\right\|_2^2
}
}.
\end{equation*}

\paragraph{\texttt{J-MLPGlobal}: Joint Backbone-LoRA MLP Global Geometry over All Prompt Tokens.}
This method is the MLP-only version of \texttt{J-Global}. 
It uses the same global cosine score, but restricts the module set to \(m\in\{\mathrm{up\_proj},\mathrm{gate\_proj},\mathrm{down\_proj}\}\):
\begin{equation*}
s_a(x)
=
\frac{
\sum_{\ell,m,t}
\left\langle
\Delta W_{a,\ell,m}h^a_{\ell,m,t}(x),
g_{\ell,m,t}(x,\hat y)
\right\rangle
}{
\sqrt{
\sum_{\ell,m,t}
\left\|
\Delta W_{a,\ell,m}h^a_{\ell,m,t}(x)
\right\|_2^2
}
\sqrt{
\sum_{\ell,m,t}
\left\|
g_{\ell,m,t}(x,\hat y)
\right\|_2^2
}
}.
\end{equation*}

\paragraph{\texttt{J-WAllZ}: Joint Weighted All-layer Geometry with Query-wise Z-score.}
This method uses the weighted all-layer dot product between the \(\mathrm{q\_proj}\) LoRA perturbation and the negative gradient, followed by query-wise z-score normalization across adapters. 
Let \(\mu_x\) and \(\sigma_x\) denote query-wise statistics over candidate adapters:
\begin{equation*}
r_a(x)
=
\sum_{\ell,t}
w_\ell
\left\langle
\Delta W_{a,\ell,q}h^a_{\ell,q,t}(x),
g_{\ell,q,t}(x,\hat y)
\right\rangle,
\qquad
s_a(x)
=
\frac{r_a(x)-\mu_x}{\sigma_x}.
\end{equation*}

\paragraph{\texttt{Arrow}: Zero-shot PEFT Routing by LoRA Parameter Geometry.}
Arrow scores each adapter by matching the dominant right singular direction of each LoRA update with the query hidden state:
\begin{equation*}
v_{a,\ell,m}
=
v_1(\Delta W_{a,\ell,m}),
\qquad
s_a(x)
=
\operatorname{Agg}_{\ell,m}
\left|
\left\langle
v_{a,\ell,m},
h^0_{\ell,m}(x)
\right\rangle
\right|.
\end{equation*}

\paragraph{\texttt{SPECTR}: Spectral Routing and Merging for Multi-LoRA Inference.}
SPECTR uses the response norm of spectral LoRA factors on token hidden states. 
Let \(\Delta W_{a,m}=B^\star_{a,m}A^\star_{a,m}\). 
The token- and module-level traces are used for top-\(k\) routing and aggregation:
\begin{equation*}
r_{a,m,t}(x)
=
\left\|
A^\star_{a,m}h^0_{m,t}(x)
\right\|_2,
\qquad
s_a(x)
=
\operatorname{Agg}_{m,t}
r_{a,m,t}(x).
\end{equation*}

\paragraph{\texttt{LAG}: Arrow Retrieval plus SPECTR Reranking for Per-token Multi-LoRA Routing.}
LAG first uses an Arrow-style direction score to retrieve candidate adapters and then applies a SPECTR-style norm for reranking. 
Let \(\mathcal C_{m,t}=\operatorname{TopK}_{a}\left|\langle a^\star_{a,m,1},h^0_{m,t}(x)\rangle\right|\) denote the retrieved candidate set:
\begin{equation*}
s_{a,m,t}(x)
=
\left\|
A^\star_{a,m}h^0_{m,t}(x)
\right\|_2,
\qquad
a\in\mathcal C_{m,t}.
\end{equation*}
The final adapter is selected by the reranking score, and selected scores are aggregated across tokens and modules.

\subsection{Existing Routers as Instances of the PMD Framework}
\label{app:pmd-reductions}

The previous section specifies the implemented baseline scores. 
Here, we provide a complementary interpretation of representative zero-shot routers under the PMD framework. 
Rather than redefining their full implementations, we identify the response object \(\rho_k(q)\) and decoder \(D\) that each method implicitly uses.
This clarifies the position of \textsc{PMDRouter}: prior routers can be viewed as using restricted, spectral, or proxy response objects, while \textsc{PMDRouter} uses the full query-conditioned linear response and decodes it with scale-normalized response energy.

\paragraph{Notation.}
Let \(h(q)\in\mathbb{R}^{d}\) denote the query-conditioned backbone activation. 
Each LoRA \(L_k\) has low-rank update \(\Delta W_k=B_kA_k\), and we write its singular value decomposition as \(\Delta W_k=\sum_{i=1}^{r}\sigma_{k,i}u_{k,i}v_{k,i}^{\top}\). 
A PMD router specifies a response object \(\rho_k(q)=\mathrm{Resp}(h(q),\theta_k)\) and a decoder \(s_k(q)=D(\rho_k(q))\).

\paragraph{Arrow.}
Arrow can be interpreted as a rank-1 PMD path. 
It keeps only the dominant right singular direction of the LoRA update and reads the corresponding projection coefficient from the query activation:
\begin{equation*}
\rho_k^{\mathrm{Arrow}}(q)
=
\sigma_{k,1}u_{k,1}\langle v_{k,1},h(q)\rangle .
\label{eq:arrow_rho}
\end{equation*}
Thus, Arrow uses a truncated response object and a decoder that retains only the top-singular projection coefficient.

\paragraph{SpectR.}
SpectR can be interpreted as a spectral PMD path. 
Instead of keeping only the top component, it aggregates multiple singular components with a fixed spectral weighting:
\begin{equation*}
\rho_k^{\mathrm{SpectR}}(q)
=
\sum_{i=1}^{r}
f(\sigma_{k,i})u_{k,i}
\langle v_{k,i},h(q)\rangle .
\label{eq:spectr_rho}
\end{equation*}
When \(f\) is the identity, this recovers the full linear response \(B_kA_kh(q)\). 
Thus, SpectR can be viewed as a spectral response constructor with a norm-style decoder.

\paragraph{LAG and SEQR.}
LAG and SEQR correspond to proxy-based PMD paths. 
LAG constructs a static adapter-side proxy \(z_k\) and scores it by \(s_k^{\mathrm{LAG}}(q)=\mathrm{sim}(\mathrm{emb}(q),z_k)\), so its response object is the query-independent proxy \(\rho_k^{\mathrm{LAG}}(q)=z_k\). 
SEQR makes this proxy partially query-conditioned by using \(\rho_k^{\mathrm{SEQR}}(q)=\tilde z_k(c(q))\) and scoring it through an external similarity function.
Both methods introduce query dependence through proxy matching rather than by directly applying the LoRA operator \(B_kA_k\) to the backbone activation \(h(q)\).

\paragraph{Position of \textsc{PMDRouter}.}
\textsc{PMDRouter} uses the full query-conditioned linear response:
\begin{equation*}
\rho_k^{\mathrm{PMDRouter}}(q)
=
B_kA_kh(q)
=
\sum_{i=1}^{r}
\sigma_{k,i}u_{k,i}
\langle v_{k,i},h(q)\rangle .
\label{eq:pmdrouter_rho_app}
\end{equation*}
It then decodes this response with scale-normalized response energy:
\begin{equation*}
D^{\mathrm{PMDRouter}}(\rho_k,q)
=
\frac{
\|\rho_k\|_2^2
}{
\|h(q)\|_2^2\cdot \|B_kA_k\|_F^2+\epsilon
}.
\label{eq:pmdrouter_decoder_app}
\end{equation*}
Compared with Arrow, PMDRouter retains all low-rank response directions. 
Compared with SpectR, it avoids hand-designed spectral reweighting and decodes the full response after scale normalization. 
Compared with LAG and SEQR, it constructs the routable object by directly applying the LoRA update to the query-conditioned backbone activation.

\paragraph{Summary and interpretation.}
Table~\ref{tab:pmd-reductions} summarizes these reductions as PMD interpretations rather than strict equivalences. 
Overall, PMD shifts routing separability from global adapter descriptors to query-conditioned responses, so globally similar LoRA memories can still be separated by their induced responses under a specific query, consistent with Figure~\ref{fig:pmd-decodability-success}.

\begin{table}[h]
\centering
\small
\renewcommand{\arraystretch}{1.18}
\setlength{\tabcolsep}{4pt}
\begin{tabularx}{\linewidth}{lXX}
\toprule
\textbf{Router} & \textbf{Response object \(\rho_k(q)\)} & \textbf{Decoder \(D(\rho_k)\)} \\
\midrule
Arrow~\citep{ostapenko2024towards}
& Rank-1 truncated linear response
& Top-singular projection coefficient \\

SpectR~\citep{fleshman2025spectr}
& Spectrally re-weighted linear response
& Norm-style aggregation \\

LAG~\citep{fleshman2025lora}
& Static or proxy adapter representation
& External proxy similarity \\

SEQR~\citep{fleshman2025seqr}
& Contextualized adapter proxy
& External proxy similarity \\

\midrule
\textbf{\textsc{PMDRouter} (ours)}
& \textbf{Full linear response} \(B_kA_kh(q)\)
& \textbf{Scale-normalized response energy} \\
\bottomrule
\end{tabularx}
\caption{
Existing zero-shot LoRA routers can be viewed as PMD instances by specifying the response object \(\rho_k(q)\) and decoder \(D\). \textsc{PMDRouter} uses the full query-conditioned linear response with a scale-normalized energy decoder.
}
\label{tab:pmd-reductions}
\end{table}

\section{Additional Method and Experimental Details}
\label{app:additional_method_experimental_details}

\subsection{Different Dataset}
\label{app: Different Dataset}

\begin{table*}[t]
    \centering
    \scriptsize
    \setlength{\tabcolsep}{3pt}
    \renewcommand{\arraystretch}{0.95}
    \resizebox{\textwidth}{!}{%
    \begin{tabular}{llcccccccccccccccc}
    \toprule
    \textbf{Method} & \textbf{Backbone} & \textbf{P01} & \textbf{P02} & \textbf{P03} & \textbf{P04} & \textbf{P05} & \textbf{P06} & \textbf{P07} & \textbf{P08} & \textbf{P09} & \textbf{P10} & \textbf{P11} & \textbf{P12} & \textbf{P13} & \textbf{P14} & \textbf{P15} & \textbf{Avg.} \\
    \midrule
    \texttt{LAG} & Llama-3.1-8B & 0.700 & 0.400 & 0.000 & 0.033 & 0.633 & 0.300 & 0.633 & 0.133 & 0.800 & 0.700 & 0.333 & 0.900 & 0.333 & 0.367 & 0.333 & 0.440 \\
    \texttt{LAG} & Qwen3-4B-Instruct & 0.367 & 0.300 & 0.400 & 0.233 & 0.767 & 0.633 & 0.667 & 0.500 & 0.800 & 0.833 & 0.500 & 0.867 & 0.667 & 0.700 & 0.733 & 0.598 \\
    \texttt{LAG} & Qwen3-8B & 0.567 & 0.400 & 0.667 & 0.233 & 0.833 & 0.667 & 0.667 & 0.567 & 1.000 & 0.833 & 0.500 & 0.733 & 0.567 & 0.667 & 0.633 & 0.636 \\
    \midrule
    \textsc{PMDRouter} & Llama-3.1-8B & 0.733 & 0.100 & 0.567 & 0.333 & 0.933 & 0.533 & 0.700 & 0.500 & 0.700 & 0.700 & 0.833 & 0.733 & 0.533 & 0.567 & 0.533 & 0.600 \\
    \textsc{PMDRouter} & Qwen3-4B-Instruct & 0.600 & 0.333 & 0.500 & 0.367 & 0.933 & 0.567 & 0.733 & 0.433 & 0.600 & 0.667 & 0.800 & 0.767 & 0.567 & 0.633 & 0.700 & 0.613 \\
    \textsc{PMDRouter} & Qwen3-8B & 0.833 & 0.233 & 0.467 & 0.433 & 0.933 & 0.533 & 0.733 & 0.533 & 0.767 & 0.733 & 0.833 & 0.767 & 0.567 & 0.733 & 0.767 & 0.658 \\
    \bottomrule
    \end{tabular}%
    }
    \caption{Per-class routing accuracy on \textsc{PaperQA}. Columns P01--P15 correspond to \texttt{paper\_01}--\texttt{paper\_15}. Both \texttt{LAG} and \textsc{PMDRouter} use the matched train-split calibration setting.}
    \label{tab:app3-paperqa-matched-per-class}
\end{table*}

\begin{table*}[t]
    \centering
    \scriptsize
    \setlength{\tabcolsep}{3pt}
    \renewcommand{\arraystretch}{0.95}
    \resizebox{\textwidth}{!}{%
    \begin{tabular}{llccccccccccccc}
    \toprule
    \textbf{Method} & \textbf{Backbone} & \textbf{Ent.} & \textbf{Hist.} & \textbf{Sports} & \textbf{Geog.} & \textbf{Music} & \textbf{Pol.} & \textbf{Lit.} & \textbf{Bus.} & \textbf{Sci.} & \textbf{Bio.} & \textbf{Tech.} & \textbf{Food} & \textbf{Avg.} \\
    \midrule
    \texttt{LAG} & Llama-3.1-8B & 0.633 & 0.893 & 0.880 & 0.480 & 0.960 & 0.587 & 0.747 & 0.627 & 0.600 & 0.827 & 0.927 & 0.640 & 0.733 \\
    \texttt{LAG} & Qwen3-4B-Instruct & 0.807 & 0.747 & 0.927 & 0.407 & 0.867 & 0.840 & 0.707 & 0.780 & 0.813 & 0.927 & 0.860 & 0.833 & 0.793 \\
    \texttt{LAG} & Qwen3-8B & 0.707 & 0.667 & 0.873 & 0.320 & 0.893 & 0.740 & 0.767 & 0.660 & 0.767 & 0.880 & 0.960 & 0.820 & 0.754 \\
    \midrule
    \textsc{PMDRouter} & Llama-3.1-8B & 0.727 & 0.800 & 0.873 & 0.533 & 0.933 & 0.627 & 0.913 & 0.653 & 0.687 & 0.940 & 0.873 & 0.673 & 0.769 \\
    \textsc{PMDRouter} & Qwen3-4B-Instruct & 0.667 & 0.847 & 0.873 & 0.480 & 0.900 & 0.720 & 0.887 & 0.727 & 0.740 & 0.967 & 0.860 & 0.707 & 0.781 \\
    \textsc{PMDRouter} & Qwen3-8B & 0.693 & 0.833 & 0.873 & 0.493 & 0.913 & 0.733 & 0.900 & 0.713 & 0.787 & 0.973 & 0.847 & 0.720 & 0.790 \\
    \bottomrule
    \end{tabular}%
    }
    \caption{Per-domain routing accuracy on \textsc{NQ-DomainLoRA}. Both \texttt{LAG} and \textsc{PMDRouter} use the matched train-split calibration setting.}
    \label{tab:app3-nq-domainlora-matched-per-class}
\end{table*}

\begin{table*}[t]
    \centering
    \scriptsize
    \setlength{\tabcolsep}{3pt}
    \renewcommand{\arraystretch}{0.95}
    \resizebox{\textwidth}{!}{%
    \begin{tabular}{llccccccccc}
    \toprule
    \textbf{Method} & \textbf{Backbone} & \textbf{BoolQ} & \textbf{Edit} & \textbf{ExtQA} & \textbf{GenCoT} & \textbf{Instr.} & \textbf{MC} & \textbf{MathCoT} & \textbf{NLI} & \textbf{Avg.} \\
    \midrule
    \texttt{LAG} & Llama-3.1-8B & 0.593 & 0.013 & 0.093 & 0.000 & 0.080 & 0.267 & 0.980 & 0.493 & 0.315 \\
    \texttt{LAG} & Qwen3-4B-Instruct & 0.067 & 0.467 & 0.087 & 0.000 & 0.647 & 0.393 & 0.860 & 0.087 & 0.326 \\
    \texttt{LAG} & Qwen3-8B & 0.293 & 0.287 & 0.000 & 0.000 & 0.320 & 0.233 & 0.607 & 0.000 & 0.217 \\
    \midrule
    \textsc{PMDRouter} & Llama-3.1-8B & 0.187 & 0.413 & 0.493 & 0.993 & 0.233 & 1.000 & 1.000 & 0.987 & 0.663 \\
    \textsc{PMDRouter} & Qwen3-4B-Instruct & 1.000 & 0.893 & 1.000 & 0.993 & 0.707 & 0.980 & 1.000 & 0.607 & 0.897 \\
    \textsc{PMDRouter} & Qwen3-8B & 0.913 & 0.740 & 1.000 & 1.000 & 0.733 & 1.000 & 1.000 & 1.000 & 0.923 \\
    \bottomrule
    \end{tabular}%
    }
    \caption{Per-task routing accuracy on \textsc{Task-LoRA}. Both \texttt{LAG} and \textsc{PMDRouter} use the matched train-split calibration setting.}
    \label{tab:app3-task-lora-matched-per-class}
\end{table*}

\subsection{Different Training Objectives}
\label{app:training_objectives}

\paragraph{Standard task loss.}
For each training example, we construct supervision only on assistant-answer tokens by masking prompt, user, and padding tokens with the ignore index \(-100\). Let \(\Omega=\{t: y_t\neq -100\}\) denote the supervised answer-token positions. The standard training objective is the causal language modeling loss:
\begin{equation*}
\mathcal{L}_{\mathrm{task}}
=
-\frac{1}{|\Omega|}
\sum_{t\in\Omega}
\log p_\Theta(y_t \mid y_{<t}, x).
\label{eq:task_ce_loss}
\end{equation*}

\paragraph{Signal-writing objective.}
When training-side signal writing is enabled, we keep the same task loss and add auxiliary losses that shape the LoRA response signals later read by the PMD decoder. All auxiliary terms are computed from frozen-backbone activations, LoRA-induced responses, or loss-derived signals, and therefore do not introduce an additional inference-time router. The overall objective is
\begin{equation*}
\mathcal{L}_{\mathrm{total}}
=
\mathcal{L}_{\mathrm{task}}
+
\lambda_{\mathrm{addr}}\mathcal{L}_{\mathrm{addr}}
+
\lambda_{\mathrm{energy}}\mathcal{L}_{\mathrm{energy}}
+
\lambda_{\mathrm{cal}}\mathcal{L}_{\mathrm{cal}}
+
\lambda_{\mathrm{ortho}}\mathcal{L}_{\mathrm{ortho}}
+
\lambda_{\mathrm{mono}}\mathcal{L}_{\mathrm{mono}}.
\label{eq:total_training_objective}
\end{equation*}

\paragraph{Addressability regularizer.}
This term encourages the low-rank addressing activation to concentrate on distinguishable coordinates. Let \(a_{\ell,k}(x)=A_{\ell,k}u(x)\) be the low-rank addressing activation, and let \(p_i(x)=a_{\ell,k,i}(x)^2/(\sum_j a_{\ell,k,j}(x)^2+\epsilon)\) be its coordinate-wise energy distribution. The entropy-based addressability loss is
\begin{equation*}
\mathcal{L}_{\mathrm{addr}}
=
-\sum_i p_i(x)\log p_i(x).
\label{eq:addr_entropy_loss}
\end{equation*}
When a top-1/top-2 separation term is used, \(p_{(1)}(x)\) and \(p_{(2)}(x)\) denote the largest two coordinates of \(p(x)\), and \(m\) denotes the target margin. The margin penalty is
\begin{equation*}
\mathcal{L}_{\mathrm{margin}}
=
\max\bigl(0, m - (p_{(1)}(x)-p_{(2)}(x))\bigr).
\label{eq:addr_margin_loss}
\end{equation*}

\paragraph{Energy contrastive loss.}
This term encourages the target memory unit to produce a stronger query-conditioned response than a competing unit. Let \(E_k(x)\) denote the response energy computed from LoRA-induced responses, let \(k^+\) and \(k^-\) denote positive and negative memory units, and let \(\tau>0\) denote the temperature. The contrastive energy loss is
\begin{equation*}
\mathcal{L}_{\mathrm{energy}}
=
-\log
\frac{
\exp(E_{k^+}(x)/\tau)
}{
\exp(E_{k^+}(x)/\tau)+\exp(E_{k^-}(x)/\tau)
}.
\label{eq:energy_contrastive_loss}
\end{equation*}

\paragraph{Energy calibration loss.}
This term explicitly calibrates positive and negative response magnitudes. Let \(c^+\) and \(c^-\) denote the target positive and negative energy levels, and let \(\beta\) control the negative-energy penalty. The calibration loss is
\begin{equation*}
\mathcal{L}_{\mathrm{cal}}
=
\max(0, c^+ - E_{k^+}(x))^2
+
\beta
\max(0, E_{k^-}(x)-c^-)^2.
\label{eq:energy_calibration_loss}
\end{equation*}

\paragraph{Response orthogonality loss.}
This term reduces overlap between the current response and an anchor or competing response. Let \(\rho_k(x)\) denote the current response, and let \(\rho_a(x)\) denote an anchor or competing response. The response orthogonality loss is
\begin{equation*}
\mathcal{L}_{\mathrm{ortho}}
=
\left(
\frac{
\langle \rho_{k}(x), \rho_{a}(x)\rangle
}{
\|\rho_{k}(x)\|_2\|\rho_{a}(x)\|_2+\epsilon
}
\right)^2.
\label{eq:response_orthogonality_loss}
\end{equation*}
A weight-space variant can be obtained by replacing the response vector with the corresponding LoRA update.

\paragraph{Monotonic alignment loss.}
This term aligns response energy with the actual improvement induced by the LoRA. Let \(I_k(x)=\ell_{\mathrm{base}}(x)-\ell_{\mathrm{lora}}(x)\) denote the loss improvement of the LoRA-augmented model over the frozen backbone, and let \(\operatorname{Corr}(\cdot,\cdot)\) denote the Pearson correlation over a minibatch. The monotonic alignment loss is
\begin{equation*}
\mathcal{L}_{\mathrm{mono}}
=
-\operatorname{Corr}\bigl(E_k(x), I_k(x)\bigr).
\label{eq:monotonic_alignment_loss}
\end{equation*}

\subsection{Different Paths of PMDRouter}
\label{app: Differnet_Paths_of_PMDRouter}

\begin{table*}[t]
    \centering
    \scriptsize
    \setlength{\tabcolsep}{3pt}
    \renewcommand{\arraystretch}{0.95}
    \resizebox{\textwidth}{!}{%
    \begin{tabular}{llccc}
    \toprule
    \textbf{Model} & \textbf{Method} & \textbf{Gold Top1} & \textbf{Oracle Top1} & \textbf{Oracle Top5} \\
    \midrule
    Qwen3-4B & GradCos & 8.89 & 8.44 & 35.11 \\
    Qwen3-4B & LogitCE & 8.67 & 8.22 & 35.33 \\
    Qwen3-4B & TopK-PseudoCE & 8.44 & 8.89 & 76.44 \\
    Qwen3-4B & Energy & 65.56 & 51.11 & 79.56 \\
    \midrule
    Qwen3-8B & GradCos & 9.33 & 9.33 & 32.89 \\
    Qwen3-8B & LogitCE & 9.33 & 8.00 & 37.78 \\
    Qwen3-8B & TopK-PseudoCE & 1.56 & 2.44 & 73.56 \\
    Qwen3-8B & Energy & 67.11 & 52.44 & 75.56 \\
    \midrule
    Llama-3.1-8B & GradCos & 9.78 & 9.33 & 34.89 \\
    Llama-3.1-8B & LogitCE & 11.33 & 8.44 & 34.22 \\
    Llama-3.1-8B & TopK-PseudoCE & 11.11 & 8.67 & 73.56 \\
    Llama-3.1-8B & Energy & 63.33 & 48.67 & 73.56 \\
    \bottomrule
    \end{tabular}%
    }
    \caption{
    Comparison of different PMD routing paths on \textsc{PaperQA}.
    Gold Top1 reports routing accuracy against the gold LoRA, while Oracle Top1 and Oracle Top5 report oracle-based selection performance under top-1 and top-5 candidate settings.
    }
    \label{tab:appendix_pmd_loss_surrogate_paths}
\end{table*}

This section summarizes several PMD routing paths used in our implementation. 
All paths follow the same PMD principle: each candidate LoRA induces a query-conditioned response object, and a decoder maps this response into a final routing score.

\paragraph{Unified notation.}
Let \(x\) denote the input query and let \(a\in\mathcal{A}\) denote a candidate LoRA memory unit. 
For layer \(\ell\), position-pooling rule \(p\), and adapted module \(m\), let the LoRA update be \(\Delta W_{a,\ell,m}=B_{a,\ell,m}A_{a,\ell,m}\). 
Let \(h_{\ell,p,m}(x)\) denote the backbone-side module input or hidden representation, where \(p\) may correspond to the assistant-prefix last token, question last token, question mean, or full-prompt mean. 
The LoRA-induced response is \(\rho_{a,\ell,p,m}(x)=B_{a,\ell,m}A_{a,\ell,m}h_{\ell,p,m}(x)\). 
When multiple modules are used, module-level scores are summed and optionally normalized by \(\operatorname{Norm}(\cdot)\).

\paragraph{PMDRouter-Energy.}
PMD-Energy decodes the LoRA-induced response by its squared Euclidean norm. 
This score measures how strongly the current query activates the parametric response of candidate LoRA \(a\). 
The final routing signal is
\begin{equation*}
S^{\mathrm{Energy}}_a(x)
=
\operatorname{Norm}
\left(
\sum_{m\in\mathcal{M}}
\left\|
B_{a,\ell,m}A_{a,\ell,m}h_{\ell,p,m}(x)
\right\|_2^2
\right).
\label{eq:pmd_energy_score}
\end{equation*}

\paragraph{PMDRouter-GradCos.}
PMD-GradCos compares the LoRA-induced response with a pseudo-loss descent direction. 
Given a pseudo answer \(y\), let \(\mathcal{L}_{\mathrm{pseudo}}(x,y)\) denote the pseudo cross-entropy loss, and let \(-\nabla_{h_{\ell,p,m}}\mathcal{L}_{\mathrm{pseudo}}(x,y)\) denote the corresponding hidden-space descent direction. 
The final routing signal is
\begin{equation*}
s^{\mathrm{GradCos}}_{a,\ell,p,m}(x)
=
\frac{
\left\langle
B_{a,\ell,m}A_{a,\ell,m}h_{\ell,p,m}(x),
-\nabla_{h_{\ell,p,m}}\mathcal{L}_{\mathrm{pseudo}}(x,y)
\right\rangle
}{
\left\|
B_{a,\ell,m}A_{a,\ell,m}h_{\ell,p,m}(x)
\right\|_2
\left\|
\nabla_{h_{\ell,p,m}}\mathcal{L}_{\mathrm{pseudo}}(x,y)
\right\|_2
+\epsilon
}.
\label{eq:pmd_gradcos_score}
\end{equation*}

\paragraph{PMDRouter-LogitCE.}
PMD-LogitCE uses a no-backward first-order approximation of pseudo-token cross entropy. 
For the first pseudo-answer token \(y\), let \(p_\theta(t\mid x)\) be the base-model next-token distribution and let \(W_t\) denote the output embedding or LM-head vector of token \(t\). 
The logit-space direction is \(W_y-\sum_t p_\theta(t\mid x)W_t\). 
The final routing signal is
\begin{equation*}
s^{\mathrm{LogitCE}}_{a,\ell,p,m}(x)
=
\left\langle
B_{a,\ell,m}A_{a,\ell,m}h_{\ell,p,m}(x),
W_y-\sum_t p_\theta(t\mid x)W_t
\right\rangle.
\label{eq:pmd_logitce_score}
\end{equation*}

\paragraph{PMDRouter-TopK-PseudoCE.}
PMD-TopK-PseudoCE uses PMD-Energy as a cheap recall stage and reranks the retained candidates by pseudo cross entropy. 
Let \(\mathcal{C}_K(x)\) denote the top-\(K\) candidate set produced by the PMD-Energy prefilter, and let \(y_{\mathrm{pseudo}}\) denote the pseudo answer used for reranking. 
The final routing decision is
\begin{equation*}
a^*(x)
=
\arg\min_{a\in\mathcal{C}_K(x)}
\operatorname{CE}
\left(
M_{\theta+\Delta_a}(x),
y_{\mathrm{pseudo}}
\right).
\label{eq:pmd_topk_selection}
\end{equation*}

\clearpage
\subsection{Judge Scores on PaperQA}
\label{app:Judge_Scores}

\begin{wraptable}{l}{0.48\columnwidth}
    \centering
    \tiny
    \renewcommand{\arraystretch}{0.58}
    \setlength{\tabcolsep}{2.6pt}
    \resizebox{0.46\columnwidth}{!}{%
    \begin{tabular}{lccc}
    \toprule
    \multirow{2}{*}{\textbf{Method}}
    & \multicolumn{3}{c}{\textbf{PaperQA}} \\
    \cmidrule(lr){2-4}
    & \makecell{\textbf{Llama}\\\textbf{3.1-8B}}
    & \makecell{\textbf{Qwen3}\\\textbf{4B}}
    & \makecell{\textbf{Qwen3}\\\textbf{8B}} \\

    \midrule
    \multicolumn{4}{c}{\textbf{Text retrieval}} \\
    \texttt{embedding} & 5.516 & 5.833 & 5.842 \\
    \texttt{bm25}      & 5.753 & 6.000 & 5.964 \\

    \midrule
    \multicolumn{4}{c}{\textbf{Backbone-only}} \\
    \texttt{B-LogitVar} & 3.696 & 3.260 & 3.331 \\
    \texttt{B-CtrL2}    & 3.727 & 3.604 & 3.304 \\
    \texttt{B-S4a}      & 3.691 & 3.469 & 3.436 \\

    \midrule
    \multicolumn{4}{c}{\textbf{LoRA-only}} \\
    \texttt{L-AlphaZ}        & 4.164 & 3.564 & 3.507 \\
    \texttt{L-HidL2-L4Z}     & 4.240 & 3.496 & 3.413 \\
    \texttt{L-HidL2-LastMed} & 3.836 & 3.231 & 3.411 \\

    \midrule
    \multicolumn{4}{c}{\textbf{Joint Backbone-LoRA}} \\
    \texttt{J-Global}    & 3.700 & 3.231 & 3.596 \\
    \texttt{J-MLPGlobal} & 3.602 & 3.189 & 3.244 \\
    \texttt{J-WAllZ}     & 3.944 & 3.096 & 3.464 \\
    \texttt{Arrow}       & 4.453 & 4.080 & 4.184 \\
    \texttt{SPECTR}      & 3.862 & 3.300 & 3.362 \\
    \texttt{LAG}         & 4.522 & 4.993 & 5.176 \\

    \midrule
    \texttt{PMDRouter}
    & \cellcolor{rankFirst}5.271
    & \cellcolor{rankFirst}5.093
    & \cellcolor{rankFirst}5.309 \\
    \bottomrule
    \end{tabular}%
    }
    \caption{
    PaperQA judge scores. \textbf{Color:} \legendbox{rankFirst}\,1st.
    }
    \label{tab:paperqa-scor-only}
\end{wraptable}

\section{Limitations}
\label{app:limitations}
This work has three main limitations. 
(1) We focus on one concrete separable response object, namely the response-energy surrogate related to estimated loss, while other PMD response objects, such as gradient alignment, output-space perturbations, or layer-specific response patterns, remain unexplored. 
(2) Although our results show that performance varies across backbones and memory granularities, we do not deeply analyze how model architecture affects document-level, domain-level knowledge, and task/skill-oriented memory routing. 
(3) PMD-Bench covers three representative EPM settings, but more scenarios, such as larger LoRA banks, multi-hop memory access, continually updated memory banks, and more diverse domain- or user-specific memory modules, should be studied in future work.

\section{Broader Impacts}
\label{app:broader_impacts}
This work studies zero-shot routing for LoRA-based external parametric memory. Its potential positive impact is to make modular LLM deployment more efficient, maintainable, and reusable by reducing the need for separately trained routing components. Potential risks include misuse of specialized adapter banks, biased or harmful behavior encoded in adapters, and reliability issues caused by incorrect routing to irrelevant memory units. Although this work is methodological and does not target high-stakes deployment, private data, or human-subject applications, practical use of LoRA memory banks should include adapter auditing, routing-failure monitoring, and appropriate access or safety controls.

\section{Cases Study}
\label{app:Cases Study}

\begin{table*}[t]
    \centering
    \scriptsize
    \setlength{\tabcolsep}{1.6pt}
    \renewcommand{\arraystretch}{1.12}
    \begin{tabularx}{\textwidth}{
        >{\raggedright\arraybackslash}p{1.55cm}
        >{\raggedright\arraybackslash}p{2.15cm}
        >{\centering\arraybackslash}p{1.05cm}
        >{\centering\arraybackslash}p{1.05cm}
        >{\centering\arraybackslash}p{1.05cm}
        >{\centering\arraybackslash}p{0.55cm}
        >{\centering\arraybackslash}p{0.55cm}
        >{\raggedright\arraybackslash}X
    }
    \toprule
    \textbf{Failure type}
    & \textbf{Diagnostic rule}
    & \textbf{Llama}
    & \textbf{Qwen3-4B}
    & \textbf{Qwen3-8B}
    & \textbf{Rank}
    & \textbf{Gap}
    & \textbf{Representative example} \\
    \midrule
        Response overlap & Gold is not separated by the matched response ranking. & 64 / 35.6\% & 82 / 47.1\% & 73 / 47.4\% & 4.68 & 5.17 & \textbf{Q:} What is the second main contribution listed in the paper?\newline \textbf{A:} The second main contribution is the introduction of three key innovations: Denoising as Tree-Rollout, Guidance Levels as Meta-Actions, and Jumpy Denoising as Fast Simulation.\newline \emph{Route:} Llama-3.1-8B; \texttt{paper\_01\_q29}; \texttt{paper\_01}$\rightarrow$\texttt{paper\_12}; rank $>5$, gap 10.0 \\
        Decoder mismatch & Gold is near the top under the matched response ranking, but is not selected. & 39 / 21.7\% & 51 / 29.3\% & 33 / 21.4\% & 2.00 & 5.26 & \textbf{Q:} What are the two data-driven approaches mentioned by name for improving the reasoning ability of LLMs?\newline \textbf{A:} RHO-1 and Phi-4.\newline \emph{Route:} Llama-3.1-8B; \texttt{paper\_02\_q02}; \texttt{paper\_02}$\rightarrow$\texttt{paper\_05}; rank 2, gap 10.0 \\
        Output near-equivalence & Wrong top-1 route, but the downstream judge gap is small. & 77 / 42.8\% & 41 / 23.6\% & 48 / 31.2\% & 3.46 & -0.62 & \textbf{Q:} What is the primary research gap identified concerning diffusion-based planners in the text?\newline \textbf{A:} The primary research gap is that it remains uncertain how diffusion-based planners can achieve 'inference-time scalability'---that is, effectively enhance planning accuracy by scaling computation at inference time.\newline \emph{Route:} Llama-3.1-8B; \texttt{paper\_01\_q22}; \texttt{paper\_01}$\rightarrow$\texttt{paper\_05}; rank 2, gap 0.0 \\
    \bottomrule
    \end{tabularx}
    \caption{
    Failure taxonomy of \textsc{PMDRouter-AE} on \textsc{PaperQA}.
    Counts denote the number/share of failures for each backbone.
    Rank is the mean gold-LoRA rank in the matched response top-\(k\) list, and Gap is the mean judge-score difference between the gold and routed LoRA.
    }
    \label{tab:pmd_failure_taxonomy}
\end{table*}


\end{document}